# Cognitive Visual-learning Environment for PostgreSQL

Manuela N. Jeyaraj, Senuri Sucharitharathna, Chathurika Senarath, Yasanthy Kanagaraj, Indraka Udayakumara
Sri Lanka Institute of Information Technology, Sri Lanka

PostgreSQL is an object-relational database (ORDBMS) that was introduced into the database community and has been avidly used for a variety of information extraction use cases. It is also known to be an advanced SQL-compliant open source Object RDBMS. However, users have not yet resolved to PostgreSQL due to the fact that it is still under the layers and the complexity of its persistent textual environment for an amateur user.Hence, there is a dire need to provide an easy environment for users to comprehend the procedure and standards with which databases are created, tables and the relationships among them, manipulating queries and their flow based on conditions in Postgresql. As such, this project identifies the dominant features offered by Postgresql, analyzes the constraints that exist in the database user community in migrating to Postgresql and based on the scope and constraints identified, develop a system that will serve as a query generation platform as well as a learning tool that will provide an interactive environment to cognitively learn PostgreSQL query building. This is achieved using a visual editor incorporating a textual editor for a well-versed user. By providing visually-draggable query components to work with, this research aims to offer a cognitive, visual and tactile environment where users can interactively learn PostgreSQL query generation.

## I. INTRODUCTION

POSTGRESQL, also simply referred to as Postgres, is an erudite open-source Object- Relational DBMS that serves as a subsidiary to practically all SQL constructs, including sub selects, transactions, and user-defined types. It possesses an established architecture that has received a strong status for dependability, data integrity, and precision. Its compatibility ranges across various operating systems. It is copiously ACID amenable, provides provision for foreign keys, joins, views, triggers, and stored procedures across omni-various languages. As stated throughout the project, the lack of awareness of Postgresql and its attached beneficiary functionalities over any other existent systems, stated the dire need to elucidate the entire workings of Postgresql. Despite its numerous benefits and functionalities, the level to which its usage has penetrated into the database community is staggering. And hence an analysis on the reasons as to why Postgresql suffers a low prominence in utilization, the constraints that keep Postgresql from being renowned and a study on existing contender systems was carried out in order to build a system that triggers learnability within a development environment for Postgresql database creation and table querying functionalities.

## II. A BACKGROUND BRIEFING OF POSTGRESQL

PostgreSQL, originally called Postgres, was created by Michael Stonebraker, a computer science professor. He started this as a follow-up of Ingres, Postgres' predecessor, which is now under the ownership of Computer Associates.
The background study on Postgres presented here was done based on the literature survey conducted on the Postgresql Documentation [1].

### A. Postgresql's support for languages and platforms
PostgreSQL is compatible with many of the dominant Operating Systems such as Windows, Linux, UNIX (AIX, BSD, HP-UX, SGI IRIX, Mac OS X, Solaris, and Tru64). It provides avid support to various forms of data that include Text, Image, Audio and Video.

It houses various programming interfaces for C / C++, Java, Perl, Python, Ruby, Tcl and Open Database Connectivity (ODBC). Postgres provisions a hefty share of the SQL standard functionalities and goes further to offer many added features as described below:

- Complex SQL queries         - Foreign Keys
- Triggers                    - Views
- SQL Sub-selects             - Transactions
- Hot Standby                 - Streaming Replication
- Multi-version concurrency control (MVCC)

PostgreSQL can also be extended to incorporate several added features in the following sections
- Data types
- Functions
- Operators
- Aggregate functions
- Index methods

In Postgres, an SQL statement is made up of tokens. These tokens can be of the following type.
- Keyword
- Identifier
- Quoted Identifier
- Constant
- Special Character Symbol

### B. Postgresql data types
Data types are specified for each column while creating tables. The advantages of doing so is that the Database tends to retain Consistency, Validity, Compactness and Performance.

*1) Consistency:* Operations against columns of the same data type give consistent results and are usually the fastest.

*2) Validation:* Proper use of data types implies format validation of data and rejection of data outside the scope of data type.
*3) Compactness:* As a column can store a single type of value, it is stored in a compact way.
*4) Performance:* Proper use of data types gives the most efficient storage of data.

The values stored can be processed quickly, which enhances the performance. Postgresql supports a wide set of Data Types. It also permits the users to create their own custom data type using CREATE TYPE SQL command. Hence the flexibility offered here is high. There is a never ending list of data types offered by Postgres.

### III. FUNCTIONALITIES RENDERED BY POSTGRESQL

*A. Creating databases in Postgresql*

There are two stagnant ways in which a user can create or define databases within the PostgreSQL environment [2].

- Using *CREATE DATABASE SQL* command
- Using *createdb* command-line executable

The parameters that could be passed in specifying the database are the **database name**, **description** and **options** such as command-line arguments which it accepts.

*B. Creating database schemas in Postgresql*

Postgresql provides the ability to create a schema where a schema is a named collection of tables that defines a structure for the database [3]. A schema can also contain **views**, **indexes**, **sequences**, **data types**, **operators**, and **functions**. Schemas are comparable to directories at the operating system level, excluding the restriction that schemas cannot be nested. PostgreSQL statement CREATE SCHEMA creates a schema [4].

*CREATE SCHEMA <name>;*

A table in the schema can be created using;

*CREATE TABLE <myschema.mytable> (...);*

*C. Postgresql Operators*

The definition of an operator explains it as an earmarked keyword or a character that bears a prime functionality in a Postgresql statement WHERE clause to accomplish the intended filter, comparison, arithmetic or related operations [5]. Operators also function as binding points that collaborate omni various statements endorsed on a query and serve the accumulated conditions' results via execution. The following operator types are currently categorized and available in Postgresql.

- Arithmetic operators
- Comparison operators
- Logical operators
- Bitwise operators

*D. Constraints and the various levels at which they can be enforced*

Constraints can be mentioned as the rubrics imposed on data columns of the database table. Constraints bar the entry of inacceptable data being passed into the database. This confirms the precision and dependability of the data in the database.

Constraints can be formulated to adhere to the column level or table level. Column level constraints are functional only on one column while table level constraints are functional to the unabridged table.

The process of outlining a data type for a column can be considered as a constraint in itself. Since data types automatically validate the values that are allowed to be stored in columns while inserting into the database, without the interference or specification from a user or developer, declaration of column data types are readily deemed as the easiest and lowest level of constraint declaration [6].

For example, a column declared with the data type *STRING* constrains the column to permit the entry of solely string values. Some of the very notably available and utilized constraints within Postgresql's operation are tabularized below.

TABLE I.

| Constraint | Description |
|---|---|
| NOT NULL | Guarantees that the column doesn't permit the entry of null values |
| UNIQUE | Guarantees that all values within a column are distinct and different from one another |
| FOREIGN Key | This constraint extends the specified constraint to the columns referenced column in another table |
| CHECK | Conditions on columns can be specified via this type of constraint. The specified constraint will be validated for all values that are stored in the column |
| EXCLUSION | This restriction certifies that if any two rows are likened on the indicated column/ columns or expression/expressions using the stated operator/operators, not all of these evaluations are bound to return TRUE |

*E. Triggers as functions in Postgresql*

PostgreSQL Triggers facilitate callback functions on the database. These are inevitably fired at the occurrence of a stated event within the or upon the database. The most commonly invoked triggers in Postgresql are listed below [7].

- Constraint is laid *before* the operation is attempted in a row (before restrictions are tested and the INSERT, UPDATE or DELETE is tried).

- Constraint is checked *after* the operation has completed (after constraints are checked and the INSERT, UPDATE, or DELETE has completed).
- Constraint is checked *instead* of the operation (in the case of an INSERT, UPDATE, or DELETE on a view).

*F. Indexes as an optimization technique*

Indexes are distinctive search techniques or tools on tables that the database search engine utilizes to fetch data quickly. This can also be stated to describe the index as a lookup or reference to the actual data in the table [8].

SELECT and WHERE clauses positively gain through the application of indexes on tables, speeding up the data retrieval procedure. But, the UPDATE and INSERT statement negatively suffer the enforcement of indexes as their execution speed is diminished. The creation or deletion of indexes bears no toll on the way in which data is present within the database. An index can be created using the following SQL statement.

*CREATE INDEX;*

This statement permits the index to have the following attributes.
- A specified name
- The table or column on which the index will function
- The order of the index (Ascending or descending)

A UNIQUE constraint can be laid upon the indexes as well. In such a way, the indexes ensure that only distinct values are entered in the columns upon which the indexes are enforced.

IV. NEED FOR A DEVELOPMENT ENVIRONMENT FOR POSTGRESQL

This proposed system is intended to induce the utilization and elucidate the importance and advantages of PostgreSQL, which is an object-relational database (ORDBMS) that was recently introduced into the database community and has been proven to be superior to the well-renown MySQL in many ways.

It is also known to be the world's most advanced SQL-compliant open source Objective RDBMS. But, users have not yet resolved to Postgresql due to the facts that it is still under the layers and the complexity of its persistent textual environment for an introductory user.

Simply stating this, there is a dire need to elucidate an easy way of making the users comprehend the procedure and standards with which databases are created, tables and the relationships among them, manipulating queries and their flow based on conditions in Postgresql to help the community resolve to Postgresql at an augmented rate.

Hence, this project tends to identify the dominant features provided by Postgresql over MySQL, analyze the constraints that exist in the database user community in moving towards the utilization of Postgresql and based on the scope and constraints identified, develop a system that will serve as a designing platform as well as a learning tool that will provide an interactive method of learning via a visual editor mode and incorporate a textual editor for a well-versed user.

By providing a visually draggable and manipulative environment to work with Postgresql databases and queries, it is expected to highlight the advanced features displayed by Postgresql over any other existent systems in order to grasp and disseminate the importance and simplicity offered by this language to a hesitant user.

*A. Concerns that led to the lack of potential users*

The reason to direct major focus towards Postgresql is the fact that in spite of its numerous advantages over any current SQL compliant DBMS, Postgresql is not a database server in vogue.

The database user community is somehow tentative to migrate to this environment abandoning conventional servers such as MySQL. As a questionable research topic on why such hesitance still exists, the following concerns were identified.
- The hassle of learning new standards and processes over well-known database management systems such as MySQL.
- Provided the advanced features of this tool, there are deficiencies in rapports of popularity, despite the extreme implementations. This affects the level of ease with which support is made available

Misconceptions about the platform that permits complex custom solutions- Postgresql is well adopted to incorporate complex custom solutions since it is extensible. Hence, based on the concerns identified, it was conclusive that learnability and comprehensibility were two major attributes that affected the user migration from other platforms to Postgresql.

So this research based system being developed tends to identify the ways and features that a development platform for Postgresql needs to have and build a development environment that will serve both the amateur users as well as the well-versed users via interactive modes of database creation and manipulation.

*B. Perspective of conjuring a feasible product*

Though there are IDEs to work on PostgreSQL databases management, these existing workspaces do not focus on inducing learnability of PostgreSQL via interactive operations. Some of the namable IDEs in this regard are PGAdmin III [9], PhpPGAdmin [10], and SQuirreL [11]. The factor considered in conjuring the solutions to differentiate the system under development from the existing systems is learnability.

Learnability of the system can be measured using the following ways.

- ***Effectiveness:*** Number of functions that could be learnt from working on the system.
- ***Efficiency:*** The time taken to learn a new function via interacting with the Application.
- ***Satisfaction:*** The perceived value of the user associated with their investment (Time, Effort and Cost) in learning how to use a particular functionality on the system.
- ***Errors:*** The number of errors that are possible to be met when working on the Application and the time or methods to recover from the incurred erroneous state.

Based on these measurable and assumable attributes, the UI design for a Visual Editor that encapsulates learnability as a key component had to be analyzed. To do so, the Info-graphic of Interaction Principles [12] by David Hogue was evaluated on a research basis. According to David Hogue's core principle of Interaction Design, Comprehensibility is coherent with learnability as shown in *Fig 1*.

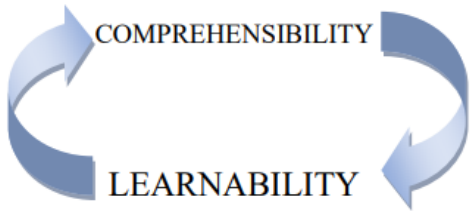

*Fig. 1*

## V. TARGET MODEL FOR POSTGRESQL LEARNABILITY

*A. System components*

Bearing in mind all the pre-calculated and evaluated features, constraints and feasible factors, a system with the following components can be considered as a viable solution to the concern at hand. The system will provide the user with functionalities under three categories.

*1) The Visual Editor*
Provides a drag-and-drop based editor where databases and queries on databases can be interactively created and manipulated.

*2) The Text Editor*
This facilitates the production of source code featuring user-programmable syntax highlighting and clause navigation functions.

*3) A user dashboard*
A personalized user dashboard for each registered user creates an illusion separating multiple logged in user sessions from one another providing privacy and manageability of one's own databases and query projects.

*B. Target users of the system*
There are basically two user classes focused within the context of this project.

*1) Novice users*
These users are those who have a certain level of database knowledge and know to work their way within a database environment in general. But their knowledge does not span well across the concepts and practicality of Postgresql in specific. Hence these users will need to be ***taught*** the concepts as well as manipulation of Postgresql Queries and databases.

*2) Expert users.*
These users are considered to be regular Postgresql users who are well-versed in its concepts and techniques. But the reason to accommodate these users within the system's context is to provide a broader scope and capture more target audience instead of extending the usability of the system to mere amateur users.

## VI. CONTRIBUTION OF A VISUAL EDITOR INCOHERENT LEARNING

The Visual Editor is predominantly intended towards the provision of learnability of Postgresql concepts and techniques for amateur users as considered within this context. As such the Visual Editor will be partitioned in two major sections based on its functionality within the system.

- Database and Table creation
- Query Builder for Table Data Retrieval

Each of these are elucidated in detail here.

*A. Database and table creation within the Visual Editor*

The various functionalities that are broken down within this major context of creating databases and tables, consider the implementation of each of these features.

*1) Effectiveness*
*a) Creating database or schema object*
Most of the web based visual editors currently provide less number of options while creating users or databases or schema objects like tables, novice users are unable to learn available options for databases or schema objects in Postgresql [13]. For example when creating a database we have to provide database configuration attributes like template, owner, collation type, character classification and connection limit. Users have to navigate pages to identify available options. So the percentage of users who manage to learn the criterion is less. Insufficient in learning a single task or function. In our tool we have provided a single form with multiple options to create database or schema objects therefore beginners may feel convenient while using, they don't need to access many pages.

*b) Continuity of task sequences*
Database design has many processes, to do each process instructions are given by our visual editor. This may be useful for both novice and expert users. For example; after logging into their accounts each user will be given a dashboard. In the dashboard we will give an instruction for users to create a new user for a new database likewise for each task we will be helping them. At a time, users can learn multiple functions.

*2) Efficiency*
Users expect to learn in a quick way. Some of the elements present in PostgreSQL are not known by beginners. They might know basic components. For each element in database design we thought of giving a small description. Postgresql provides many data types but some data types are unknown by novice users. For instance as shown in *Fig. 2*, while creating a table with columns and when the user moves the cursor to that particular data type like double precision, serial our visual editor describes it. So users no need to get help from other sites. This feature may reduce time consumption to learn and is a convenient way to increase their knowledge on Postgresql.

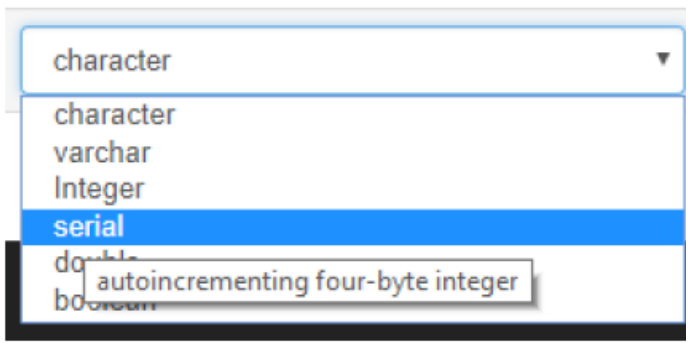

*Fig. 2*

Database beginners may be confused if they see numerous options in a single form of table creation. We can't limit the options or fields because the system should be used by experts simultaneously. Therefore to make it more efficient we separated table options into two parts as shown in above *Fig. 3*.

- Primary details which includes table name & description
- Advance properties which include indexes, check constraints [14] and schemas.

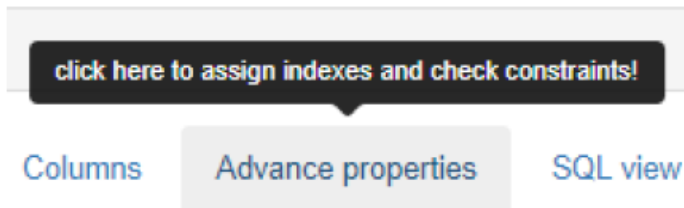

*Fig. 3*

*3) Satisfaction*

Satisfaction can be measured using a rating scale of ease of learning. This factor includes effort, time and cost. Main effort of users is learning the Postgresql database system. To increase the learnability, the number of components in UI must be reduced [12]. Current visual editors have so many menus after logging in. By providing simple forms for users and database creation, beginners might adapt to the system for the learning process. In the user creation task of our product, the available components are user name and single or multiple user configuration selections like super user, created role, and password [15].

*a) Visibility of commands and menu options*

Make commands and menu options highly visible and easy to find. For example near the database object that they impact - a right click on that database or if they click on the dropdown icon displays a list of available operations like creating new table and view existing tables as shown in below *Fig. 4*.

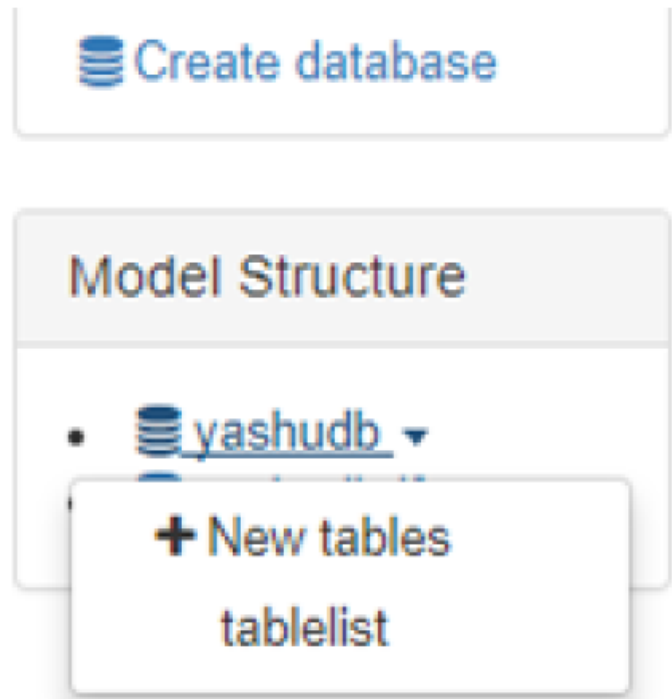

*Fig. 4*

*b) Providing command feedbacks*

Prompting out successful or failure messages after object creation this feature may help users to find out whether their schema objects were created on their specific database. As shown in *Fig.3* if a user clicks on '+ New table', it redirects to create a table page. After creating the table it shows a successful or failure message. According to time satisfaction, if they navigate the table list, users will be able to view existing tables of that database, add columns menu and table drop option. They don't need search or navigate multiple pages to choose options like drop table menu

*4) Recover from errors*

In Postgresql certain index methods do not support some data types of columns [16]. If a user selects incompatible data types while selecting index methods, the system should give an error message for the user to discover the error. Unique option is compatible with some index methods of tables like 'btree' index method supports unique option but 'hash' method does not have unique option. So the solution for the above problems is making the unique field visible only when the user selects certain index types.

These validation avoid operations that do not succeed because of some simple and predictable mistake. Providing guidance if an inappropriate command is activated or enable commands only when they can be used in the correct context [17]. For instance if a user starts typing database names in digits, we should provide an error message and give a solution to sort it out like giving a message to start with letters.

*B. Query Builder for table data retrieval within the Visual Editor*

The query builder for Postgresql table queries is an interactive environment that consists of the following 3 elements.

- The Toolbox
- The Canvas
- Properties panel

The Toolbox will be a host of the elements that represent the various clauses in the Postgresql query statements. Users will be permitted to drag and drop the elements onto the canvas. Preceding the drop of an element onto the canvas, that element will be cloned on the canvas so that the user can freely move and manipulate this element within the canvas. The canvas will contain the elements' movement so that the elements are not extendable out of that container. The properties panel is a dynamic property declaring window that is created and made visible only when the user chooses to set the properties of a particular element. Based on the element chosen, the property attributes of the element will be gathered with regard to the connections that the element bears at the moment within the container and display the available and valid properties that could be assigned for that element.

The crucial features that were considered in formulating the features, design principles, location, minimization and visibility of the interface elements of the Query Builder Interface in order to facilitate learnability are stated here.

As suggested via Hogue's core principles, there needs to be Evidence of action. Learnability is provided through traceability of a certain action. In touch screen functionalities, the previous actions are lost unless they are stored by coded means. Likewise, when an application's intention is to provide a learning platform, the users' previous actions need to be traceable to identify and correlate the tasks performed with the output rendered.

To solve this issue, the Visual editor will provide a "*History*" option where past actions or functions performed can be viewed by the user as part of the model and its connected network of elements that represent the query clauses. In addition, a user is found to spend more time learning a feature if that feature bears the following characteristics.

- *The presence of a mandated feature that they are bound to use often*

Highlight the clauses and functional elements that form the basis of any PostgreSQL statement by means of instructions and color pop-ups.

- *Provides alternatives*

Once the user completes a query creation, suggest alternative query optimization patterns to enhance a cost-effective design.

- *Is attained at low cost*

The system will be implemented as a web application incurring no additional software or hardware purchases.

- *Provides a simple first interaction*

Guide new users via tooltips, suggestions and instructions while permitting the option to limit an overload of instructions at the same time.

In balancing learning components with UI components, the proportion of UI features should maintain a minimalistic design without overwhelming the users' cognitive load [18]. Hence the Visual Editor will adhere to a minimalistic, simple design where only the basic elements will be displayed initially. Upon the users' will, advanced features of PostgreSQL can be accessed and utilized at no cost. A balance between learnability and UI component features needs to be optimized to provide an environment that facilitates teaching in subtlety without letting the user be disrupted as shown in *Fig. 5*.

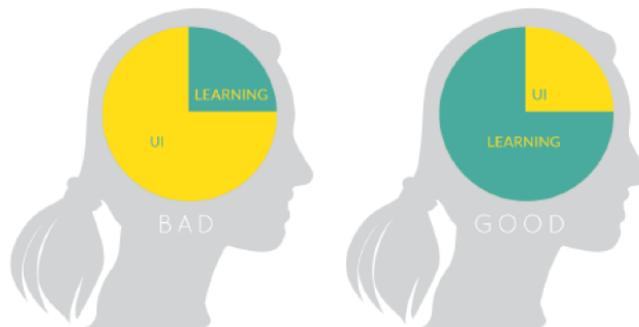

*Fig. 5*

The following features will be implemented to make learning easy via the usage of the application.

*a) Scaffolding*

This is the process where the application will direct an amateur user to gradually move into the difficulty level via properly guided instructions and tailored clause flows.

*b) Quiet Design*

The UI will be designed to silently lay in the background to avoid overwhelming the user with tons of components/elements or functionalities all at once.

*c) Paradox of Choices*

Eliminating consumer choices can greatly reduce anxiety for shoppers and reveals a global disconnect between happiness and freedom [19]. Based on this revelation by Barry Schwartz, more options does not directly indicate user freedom. Instead it toils with happiness increasing anxiety.
Applying this concept to our Visual Editor, providing the basic and mandatory elements at first sight while hiding the advanced options will provide a limited operational environment. But this environment will allow the user to get accustomed to the available functions before moving forth into the advanced option.

Once the user familiarizes with the limited base set of functionalities, the advanced options can be accessed to proceed further on.

*d) Mental Models*

This is what the user expects the system to perform. There are 3 ways to bind the users' mental model with the actual system.

Redesign the system to conform to users' understanding.
Design UI to better communicate the system's nature in order to correct the users' mental model.
Educate users to adapt to the application environment.

Out of the afore-mentioned, option b will be followed by conducting onsite observation, discussion and interviews with potential user classes to draw up an idea about the users' mental model. Based on the conjured mental model, the UI will be designed to align with the users' expectations.

When option b fails and the user is left with no clue of the next step to proceed onto, the user can choose to be educated via a brief tutorial that elucidates the system's functionalities from scratch.

*e) Intrinsic Complexity*

This is about reducing the user's extraneous cognitive load [12] by making each element's functionality more obvious to the users to avoid making them remember it.

*f) Expertise Reversal Effect*

Experts do not need information to be chunked as they can handle more control over their instruction. But at the same time these experts won't be hurt by design guidelines that apply to the novice users.

Though this is the case, sometimes rules for novices can decrease the learning outcomes for experts by slowing them down. So since our target group comprises both the novice and expert users the instructions and guidelines will be tailored to provide enough information for a novice user to work on the system and simultaneously not hinder an expert user's work effort/speed.

*g) Desirable level of Difficulties and Errors*

Errors need not to be necessarily seen as an undesirable effect. Instead errors pave the way to learn through experience. Hence, the application will provide a level of guidance that permits the user to falter occasionally. But methods of recovery and guidance will be available upon the user's breaking point.

Since the other existing products with relation to PostgreSQL do not focus on providing a learning platform for users as well as an operational environment for experts, the system under development in our context, based on all of these conclusive learnability features, the Visual Editor's UI design and component features have been drawn for implementation.

*1) Mapping the elements to the query clause representation*

The query builder's intention is to create a visually comprehensible model that educates the user about the query flow and execution by simply being able to display it obviously.

TABLE II.

| Postgresql Query Clause/function | Visual Editor Representation | |
|---|---|---|
| | *Name* | *Element Representation* |
| SELECT | SELECT | 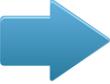 |
| FROM | TABLE | 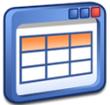 |
| WHERE | WHERE | 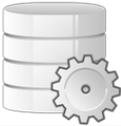 |
| GROUP BY | GROUP BY | 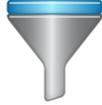 |
| HAVING | HAVING | 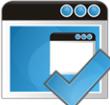 |
| ORDER BY | ORDER BY | 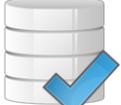 |
| Join | JOIN | 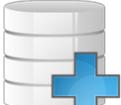 |

As such, the chosen elements that are called to map the clauses need to adhere to the meaning of the clause directly or indirectly. Hence, the following elements were chosen to represent the query clauses for easy comprehensibility.

Likewise, connectivity between dropped elements, the properties panel elements and the dynamicity in the overall flow and interface components of the visual editor are all aligned towards the provision of an exclusive query builder that not only functions as a development environment, but also a learning environment.

VII. THE CONTRIBUTION OF A TEXT EDITOR WITHIN THE PROPOSED SYSTEM

The Text editor is a mere execution text pad that paves way for the Postgresql database user to type in text, execute and run it in order to retrieve intended results or queries on the database. This scenario can be atomized into the following functionalities.

- The Execution Environment
- The Enhanced provident Environment

These are an encapsulation of the two major functionalities that the Text Editor will work with. As such each of these wide classes are elucidated in detail for clarity by providing the exact mannerism of the Text Editor's trivial tasks.

*A. The Execution Environment within the Text Editor*

When it comes to quality of a software system it mainly focuses on "capability of a software product to conform requirements" (ISO/IEC 9001). And also this can be defined as customer value (Highsmith, 2000). All these definitions are guidelines for one main requirement of a software system. This illustrates how the software system architecture covers the stakeholders non-communicate requirements. In there it is important to understand what the main objective of using the software is.

If the Software system only addresses the stakeholders main verbally communicated requirements which are considered as functional requirements it will not be considered as a customer valued software system. Software system quality represents one or more structural aspects, which illustrates how the architecture addresses the concerns such as requirements, objectives, intention of stakeholders for the architecture design [20].

*1) The usability aspect of Postgresql Text Editor*

Usability of a software system is one of the main quality attributes. It focuses on the ability of a system to perform well. For a given task how the software system performs. How the system performance can be improved. Usability of a system addresses a user who is a novice user. Assuming that the user does not have experience with computers, Experience with interface is none, domain knowledge is none and experience with similar type of software is none [21].

Usability is the ease of use and the learnability of a human made object such as a tool or device. In Software Engineering usability is the degree to which a software can be used by specific consumers to archive and quantify objectives with effectiveness, efficiency and satisfaction in a quantified context of use. Most of the software systems are rejected by the community because of this usability factor is lacking from those systems. Usability is simply the usefulness and the end user satisfaction.

*2) Learnability via the Text Editor*

Learnability is a main aspect of usability. In order to improve usability of a software system it is important to improve its characteristics. Such as usefulness, ease-of-use, end user satisfaction and learnability. But still there is little agreement on improving the learnability factor. Within the past few decades there were lots of methodologies matured to provide well accepted usability of a system. But still there is a little agreement for improving learnability. In a later study Lazer et al found that the users lose up to 40% of their time due to "frustrating experience" with computers. With one of the most common causes of this frustration being missing, hard to find and unusable features of the software system [22].Learnability is not only false under HCI literature. It is also false under fields such as artificial intelligence [23], CSCW [24], Psychology [25] and technical communication [26]. Therefore the designer who is responsible to improve the learnability has to be aware of those research areas as well.

Even though the main aspect of usability still remains, it will bear a problem in defining learnability. When developing a system interface the most difficult part is to implement an ease-of-use feature. When the user is working on the system for the first time, he may need support or a guide to perform well. To provide that feature within the given environment is a difficult part when there is no definition or a standard. Reason for this is because learnability is not a term evolving from a particular research area. It has evolved in many research areas. In 1980 Michelson et al defined learnability as the system should be easy to learn by The class of intended users for whom it is intended to [27].This definition is a positive effect when optimizing difficulties when implementing learnability features. Nielsen defines learnability as novice user experience of the initial part of the learning curve [28].However high learnable system can be considered within a short time [21].

*3) Methodology in implementing learnability*

Human mind tends to respond to visual representation rather than textual representation. Therefore, learnability can be enhanced by using graphical representation. Which means if the system can guide the user system then that system will be a better guide for the user. Every system has a support option which acts as a manual guide to the user. But since of the complexity of that manual user will not tend to use it. Because there can be more systems that can be used for the same intention with easy procedures to follow. Therefore supportive function cannot be considered as an effective fact which can be used to enhance the learnability of the system. It can be helpful for a user who has the knowledge of using similar types of systems.

*4) Learnability methodologies within the Postgresql IDE – Text Editor context*

Postgresql IDE will guide the user through out of the process using visual representations as well as textual representation. Rather than providing comprehensive manual documents it has a better way of improving the learnability of the system. Postgresql editor will help or will guide the user throughout the procedures of query processing. It will use interactive popup windows and models to attract the user and help to develop queries. For a novice user this guidance will provide better experience in learning PostgreSQL.

If users do not have the knowledge of PostgreSQL querying then they only have to type the required pseudo code on the text editor. And the query will be automatically generated from the system. System will identify the query and it will generate the query automatically after reading relevant keywords.

Query auto completion is another feature that will be available from this system. Users only want to add half the query so that the whole query will get auto completed. System will identify the relevant keyword and complete the whole query with the usage of these keywords.

Error reduction is also included for this IDE. Where the system will identify the errors and correct them at the point of identification. And also it will provide a better description of the error so that the user can learn about the source of the error. Error description will be in more readable language without having more technical words that make the description more comprehensive to the novice user.

By using these methodologies text editors will provide a better learning experience to the user in the context of PostgreSQL. And also these features will guide the user to learn the system within a short time.

*B. The Enhanced provident Environment within the Text Editor*

Enhancement of a user provided query is available through an optimization option that the text editor will provide. The query optimization intends on suggesting alternatives to optimize the existing query that the user has typed.

*1) Query Optimization via the Text Editor*

Query optimization plays a vital role in database management systems. The best way to tune performance is to try to write queries in a number of different ways and compare their reads and execution plans [29]. Given a query, there are many plans that a database management system (DBMS) can follow to process it and produce its answer. All plans are equivalent in terms of their final output but vary in their cost. Various techniques can be used to optimize database queries in order to improve the performance of PostgreSQL queries [30]. Therefore query optimization is absolutely necessary in a DBMS. The cost difference between two alternatives can be enormous. For example, consider the following database Postgresql query [31]. This example below shows the same query with two different query plans.

```
SELECT SSN, address
FROM Customers
WHERE credit_score (SSN) >600
AND education='College';
```

For the query shown above, the query plan can be drawn in two different ways. The first manner is shown in *Fig. 6*.

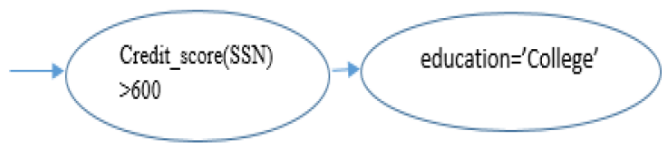

*Fig. 6*

The second alternative plan for the query considered within this scenario is shown in *Fig. 7*.

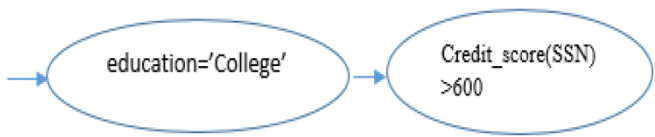

*Fig. 7*

The proposed Postgresql IDE is a designing and learning tool. Therefore increasing the learnability about query optimization is the objective of this Postgresql text editor. This tool provides a way for users to get a clear idea about query optimization. Users can get the query plan of executed Postgresql query. It contains query execution time, query planning time, query cost and other details. Output displays in a user friendly way. Therefore it will increase the learnability and the usability of the tool. This tool provides query optimization tips for each and every query. It will help users to get an idea about whether the query is further optimizable or not. Users need to understand the query optimization tips and change their query accordingly and observe the execution time. By comparing query plans users can decide which query is the best for the execution. As this tool fulfills the ultimate goal of any database system is to allow efficient querying by minimizing response time of the queries through performance tuning.

Some techniques that were adhered in order to facilitate query optimization are stated here below [32].

*a) Specific column name instead of * in SELECT*
*Example:* Writing the query as
```
SELECT col_1, col_2, col_3, col_4
FROM table_name;
```
Instead of:
```
SELECT * FROM table_name;
```

*b) Try to avoid HAVING Clause in Select statements*
HAVING clause is used to filter the rows after all the rows are selected and is used like a filter. Hence the HAVING clause should be avoided for any other purpose.

*c) Avoid the use of DISTINCT clause where applicable*
DISTINCT clauses will result in performance degradation. So this clause should be used only when it is necessary or unavoidable.

*d) Alternatives of COUNT (*) for returning total tables row count*
If the table's row count is required to be returned, alternative ways instead of the `SELECT COUNT (*)` statements need to be used. As `SELECT COUNT (*)` statement makes a full table scan to return the table's row count, it can take much time for the large tables.

*e) Minimize number of Sub query blocks within query*
There may be more than one subquery nested within the main query. Hence the number of sub query blocks in the main query need to be reduced. So the query needs to be written as shown here.

```
SELECT col1
FROM tablename1
WHERE (col2, col3) =
(SELECT MAX (col2), MAX (col3)
FROM tablename2)
AND col4 = testvalue1
```
Instead of:

```
SELECT col1
FROM tablename1
WHERE col2 =
(SELECT MAX (col2)
FROM tablename2)
AND col3 =
(SELECT MAX (col3)
FROM tablename2)
AND col4 = testvalue1;
```

*f) Using UNION ALL instead of UNION*

The UNION ALL statement is faster than UNION, because UNION ALL statement does not consider duplicates.

CONCLUSION

Postgresql is an Object Relational Database Management System that has received diminished reputation and utilization among the database community over the years. In-spite of its extensive list of support and functional features over its competitors, Postgresql is still not visibly renowned. Hence, in order to create more awareness and migration towards Postgresql, the current issues and concerns that relate to this low level of Postgresql users had to be identified. As such, the current features provided by Postgresql, the dominant features that Postgresql possesses over its competitors, the constraints that it suffers as a product in the market were analyzed and based on that a viable system has been proposed and developed. This system will function as a development environment for Postgresql databases and queries. But due to the reasons identified via the factors under concern, the lack of Postgresql knowledge among users dictated the system under development to bear learnability as a mandated feature in its environment. Hence, the system has been developed by researching the provision of learnability through visual manipulations and interactivity via a Visual Editor and for the more advanced users, a Text editor with query optimization and execution facilitations. This paper tends to revolve around the simultaneous provision of learnability and a development environment to both novice and professional users through Visual Elements and interactive dynamicity.

REFERENCES

[1] Official PostgreSQL Website. (n.d.). Retrieved March 12, 2017, from https://www.postgresql.org/

[2] Point, T. (2017, July 23). PostgreSQL - Database Creation. Retrieved August 13, 2017, from https://www.tutorialspoint.com/postgresql/postgresql_create_database.htm

[3] Definition of a database schema. (n.d.). Retrieved August 13, 2017, from http://searchsqlserver.techtarget.com/definition/schema

[4] Database schema- A breakdown of its concepts. (n.d.). Retrieved August 13, 2017, from http://workshops.boundlessgeo.com/postgis-intro/schemas.html

[5] PostgreSQL - Operators. (2017, July 23). Retrieved August 13, 2017, from https://www.tutorialspoint.com/postgresql/postgresql_operators.htm

[6] Constraints. (n.d.). Retrieved August 13, 2017, from https://www.postgresql.org/docs/9.4/static/ddl-constraints.html.

[7] Postgresql Triggers. (n.d.). Retrieved August 13, 2017, from https://www2.navicat.com/manual/online_manual/en/navicat/linux_manual/TriggersOthersPGSQL.html

[8] SQL Indexes. (2017, July 23). Retrieved August 14, 2017, from https://www.tutorialspoint.com/sql/sql-indexes.htm.

[9] PgAdmin III . (n.d.). Retrieved April 21, 2017, from http://www.pgadmin.org/

[10] PhpPgAdmin. (n.d.). Retrieved April 21, 2017, from http://sourceforge.net/projects/phppgadmin

[11] SQuirreL. (n.d.). Retrieved April 21, 2017, from http://squirrel-sql.sourceforge.net/

[12] Learnability: 5 Principles of Interaction Design to supercharge your UI (3 of 5). (2017, March 06). Retrieved April 23, 2017, from http://www.givegoodux.com/learnability-5-principles-interactiondesign-supercharge-ui-3-5/

[13] Create a Database. (n.d.) Retrieved August 16, 2017, from https://www.postgresql.org/docs/9.0/static/sql-createdatabase.html

[14] 5.3. Constraints. (n.d.). Retrieved April 23, 2017, from https://www.postgresql.org/docs/8.1/static/ddl-constraints.html

[15] CREATE USER. (n.d.). Retrieved August 16, 2017, from https://www.postgresql.org/docs/9.5/static/sql-createuser.html

[16] Index Types. (n.d.). Retrieved April 27, 2017, from https://www.postgresql.org/docs/9.1/static/indexes-types.html

[17] Nielsen Norman Group. (n.d.). Retrieved August 16, 2017, from https://www.nngroup.com/articles/usability-101-introduction-to-usability/

[18] Van Gerven, Pascal W. M. (2003/03/01). Cognitive Load Measurement as a Means to Advance Cognitive Load Theory. Educational Psychologist, 38, 63-71. doi: 10.1207/S15326985EP3801_8

[19] Schwartz, B. (2004, January). The paradox of choice: Why more is less. New York: Ecco.

[20] Dr.A. Chandrasekar, Mrs Sudharajesh, Mr P Rajesh. A Research Study on Software Quality attributes. International Journal Of Scientific & Technology Research Volume4, Issue 1, January 2014 ISSN 2250-3153

[21] Tovis Grossman, George Fitmaurice, Ramtin Attar. Autodesk Research. 210 King St.East, Toronto,Ontario, Canada, M5A 1J7

[22] Lazar, J., Jones, A. and Shneiderman, B. (2006). Workplace user frustration with computers: An exploratory investigation of the causes and severity. Behaviour and Info. Technology. 25(3):239-251

[23] Howes, A. and Young, R. M. (1991). Predicting the learnability of task-action mappings. ACM CHI. 1204-1209.

[24] Twidale, M. B. (2005). Over the Shoulder Learning: Supporting Brief Informal Learning. CSCW.14(6):505-547.

[25] Stickel, C., Fink, J. and Holzinger, A. (2007). Enhancing Universal Access–EEG Based Learnability Assessment. Lecture Notes in Comp. Sci. 813-822.

[26] Haramundanis, K. (2001). Learnability in information design. ACM SIGDOC. 7-11.

[27] Michelsen, C. D., Dominick, W. D. and Urban, J. E. (1980). A methodology for the objective evaluation of the user/system interfaces ofthe MADAM system using software engineering principles. ACM Southeast Regional Conference. 103-109.

[28] Nielsen, J. (1994). Usability Engineering. Morgan Kaufmann.

[29] P.Jyotsnai, P.Sunil Kumar Reddy, P.Govindarajulu, Dept of Computer science,S.V.University, Tirupat(2013) Effective Implementation Of Query Optimization Through Performance Tuning Tecniques On Web

[30] Yannis E. Ioannidis,Query Optimization,Computer Sciences Department,University of Wisconsin,Madison, WI 53706

[31] Saurabh gupta,Gopal Singh Tandel,Umashankar Pandey, A Survey on Query Processing and Optimization in Relational Database Management System

[32] Jean Habimana, Query Optimization Techniques - Tips For Writing Efficient And Faster SQL Queries,International Journal Of Scientific & Technology.